\documentclass[letterpaper]{article} 
\usepackage{aaai2026}    

\usepackage{times}                   
\usepackage{helvet}                  
\usepackage{courier}                 
\usepackage[hyphens]{url}            
\usepackage{graphicx}                
\usepackage{natbib}                  
\usepackage{caption}                 
\usepackage{algorithm}
\usepackage{algorithmic}

\usepackage{amsmath}
\usepackage{amssymb}
\usepackage{amsfonts}
\usepackage{bm}
\usepackage{booktabs}
\usepackage{multirow}

\newcommand{\R}{\mathbb{R}}

\newcommand{\N}{\mathcal{N}}

\newcommand{\spdom}{\Omega}              
\newcommand{\rrr}{\mathbf{r}}            

\newcommand{\hcore}{g}                   
\newcommand{\Wcore}{G}                   

\newcommand{\TODO}[1]{}

\title{HarmoCore: Functional Latent Diffusion for Sparse Reconstruction of Oscillatory Wave Fields}

\author{
    Lihao Chen$^1$, Xinyu Zhang$^1$, Panqi Chen$^1$, Lei Cheng$^1$, Ting Zhang$^1$, Jianlong Li$^1$, Shikai Fang$^1$\thanks{Corresponding author: Shikai Fang $<${\texttt{fsk@zju.edu.cn}}$>$}
}
\affiliations{
    $^1$College of Information Science and Electronic Engineering, Zhejiang University
}

\begin{document}
\maketitle


\begin{abstract}
Reconstructing oscillatory wave fields from scattered sensors is a
severely underdetermined inverse problem. Beyond the challenges of
general physical-field reconstruction, wave responses are
complex-valued, frequency-sensitive, and highly oscillatory, while
costly simulation and sensing often leave only extreme-sparse
observations. Existing low-rank, operator, and diffusion approaches are
largely designed for real-valued, smoother fields; dense pixel-space
diffusion is particularly inefficient for oscillatory complex fields
and difficult to scale to 3D. We propose \textbf{HarmoCore}, which
places a generative prior in a compact, continuous, and structured
wave-field latent. HarmoCore represents joint real--imaginary channels
with Functional Tucker cores over shared continuous spatial bases,
learns a frequency-conditioned core diffusion prior, and performs
Diffusion Posterior Sampling directly in core space. At fixed sensor
coordinates, the multilinear decoder induces an explicit likelihood
guidance operator, avoiding dense pixel-space correction. Optional
target-equation residual guidance further promotes physical
consistency. Experiments on 2D Helmholtz, 2D synthetic wave fields, and 3D
Helmholtz show substantial gains under $1\%$--$2\%$ sensing while
remaining practical in three dimensions.
\end{abstract}



\section{Introduction}
\label{sec:intro}

Reconstructing physical fields from sparse measurements is a
fundamental inverse problem across science and
engineering~\cite{arridge2019inverse,manohar2018sensorplacement}. Wave-field
reconstruction is particularly important in electromagnetic
simulation~\cite{colton2013inverse}, ocean
acoustics~\cite{jensen2011ocean}, geophysical
imaging~\cite{virieux2009fwi}, and many other
sensing and modeling problems. We focus on time-harmonic wave fields,
which describe the steady-state response to excitation at a fixed
frequency. In these applications, high-fidelity
simulation or dense sensing can be prohibitively expensive, especially
when responses must be acquired across many frequencies. The available
training fields and sensor measurements are therefore often severely
limited. Recovering a complete wave response from scattered sensors is
thus both practically important and profoundly underdetermined.

Existing sparse-field reconstruction methods broadly include
low-rank fitting, deep sparse-to-dense regression, neural operators,
and diffusion-based generative reconstruction. Low-rank methods exploit
compact spatial structure~\cite{newman2021functional,luo2024lrtfr};
neural networks
and operators learn direct field mappings~\cite{van2021voronoi,li2021fno,tran2023ffno,lu2021deeponet,li2021pino,li2023geofno,kovachki2023no};
and diffusion priors can be combined with partial observations through
posterior sampling~\cite{chung2023diffusion,gao2024diffusionpde}.
These paradigms have shown strong results, but most have been developed
for real-valued, relatively smooth spatial or spatiotemporal fields.
Sparse reconstruction of frequency-sensitive, highly oscillatory
complex-valued wave fields remains largely underexplored.

Oscillatory wave fields, however, differ sharply from the smoother data
targeted by most existing methods. In the time-harmonic setting, the
Helmholtz, frequency-domain Maxwell, and elastodynamic equations define
a complex spatial response at each frequency. Its real
and imaginary components jointly encode amplitude and phase, while
sources, materials, and boundaries create nonlocal interference. Short
wavelengths and sensitivity to frequency or medium parameters produce
rapid spatial variation and can shift nodes and antinodes throughout
the domain, leaving local sensors weakly informative about unobserved
regions. Reconstruction is therefore not merely local interpolation:
the model must infer a globally coherent phase pattern from sparse
evidence, and small phase errors can alter interference across the
domain. This structure mismatches common representations. Fixed
low-rank models can suppress frequency-dependent modes; learned
regressors and operators require broad training coverage to distinguish
phase-sensitive responses; and pixel-space diffusion must model every
rapidly varying complex value, making guidance costly and 3D scaling
difficult. Generic visual latents remain grid-bound and are not designed
for continuous-coordinate spatial queries or coupled complex channels.
The central challenge is therefore to build a generative prior aligned
with both the oscillatory field structure and its sparse observations.

To address these challenges, we propose \textbf{HarmoCore}, a functional
latent diffusion framework for sparse complex wave-field
reconstruction. We represent the real and imaginary components of each
field as joint channels of a compact Functional Tucker core over shared
continuous spatial bases. The bases capture continuous coordinate
dependence, while the core retains compact, sample- and
frequency-specific coefficients. We train a frequency-conditioned
diffusion model on these cores and perform Diffusion Posterior Sampling
directly in core space. We evaluate the shared bases at the sensor
coordinates, so we can use scattered observations without
rasterization. We exploit the decoder's multilinearity to write the
sensor measurements as a linear operator on the core, which gives a
closed-form observation-likelihood gradient. For optional
governing-equation guidance, the same decoder provides a fixed
core-to-field Jacobian that efficiently propagates the gradient of a
possibly nonlinear residual. We thus keep posterior correction in the
compact core space instead of the dense pixel space. Experiments on 2D
Helmholtz, 2D synthetic wave fields, and 3D Helmholtz demonstrate substantial
gains under $1\%$--$2\%$ sensing.

We summarize our contributions as follows:
\begin{itemize}
    \item We propose a frequency-aware, compact representation of
    complex oscillatory wave fields based on Functional Tucker models,
    with joint real--imaginary channels and continuous-coordinate
    decoding.
    \item We train a frequency-conditioned diffusion prior in the core
    space and exploit the multilinear decoder to obtain a closed-form
    observation-likelihood gradient and efficient governing-equation
    residual guidance.
    \item We demonstrate substantial improvements under extreme
    sparsity across 2D Helmholtz, 2D synthetic wave fields, and 3D Helmholtz
    reconstruction settings.
\end{itemize}


\section{Preliminaries and Problem Setup}
\label{sec:prelim}

\subsection{Sparse Reconstruction of Time-Harmonic Wave Fields}
\label{sec:prelim:task}

A time-harmonic wave field describes the steady-state complex spatial
response to excitation at angular frequency~$\omega$.
The physical field $\tilde{u}_s(\rrr,\omega)\in\mathbb{C}^K$
encodes amplitude and phase jointly; real and imaginary parts together
determine interference, node locations, and energy distribution
throughout the domain.
We represent the field through stacked real and imaginary channels
and formalize the sparse observation model as
\begin{equation}
\begin{aligned}
u_s(\rrr,\omega)
  &= \bigl[\operatorname{Re}(\tilde{u}_s),\,
           \operatorname{Im}(\tilde{u}_s)\bigr]
   \in \R^C,\quad C = 2K,\\
\mathbf{y}_m
  &= u_s(\rrr_m,\omega) + \varepsilon_m,
   \quad \varepsilon_m \sim \N(\mathbf{0},\sigma_{\mathrm{obs}}^2 I),
\end{aligned}
\label{eq:field_obs}
\end{equation}
where $s$ indexes the field instance (source location, material
parameters, boundary configuration), $\rrr_m\in\spdom$ is the
$m$-th sensor coordinate, and $\mathbf{y}_m\in\R^C$ is the measured
channel vector.
For scalar wave fields $K=1$ ($C=2$, one real and one imaginary
channel); in the experiments reported here all three benchmarks
use $K=1$.
The sparse observation set $\{(\rrr_m,\mathbf{y}_m)\}_{m=1}^M$
may cover only a small fraction of the evaluation grid; in our
extreme-sparse experiments, for example, the sensing ratio is as low
as $1$\%--$2$\%.
The primary reconstruction target is the full continuous field
$u_s(\cdot,\omega)$ over $\spdom$, evaluated on both observed and
unobserved regions, with the unobserved region being the main
indicator of reconstruction quality.

The fields satisfy a governing equation
$\mathcal{L}_{\xi_s,\omega}\,\tilde{u}_s = f_{\xi_s,\omega}$, where
$\mathcal{L}$ is the differential operator (e.g., the Helmholtz
operator $-\Delta - \omega^2/c(\rrr)^2$), $\xi_s$ collects the
instance-specific medium parameters, and $f_{\xi_s,\omega}$ is the
source term.
Operator and source metadata available at test time are
dataset-specific and used for optional equation guidance in
Section~\ref{sec:method:dps}.

\subsection{Functional Tucker Representations}
\label{sec:prelim:ftm}

Tucker decomposition approximates a tensor
$\mathcal{X}\in\R^{n_1\times\cdots\times n_d}$ with a compact core
$G\in\R^{r_1\times\cdots\times r_d}$ and factor matrices
$U_k\in\R^{n_k\times r_k}$, $r_k\ll n_k$, reducing the parameter
count from $\prod n_k$ to $\prod r_k + \sum n_k r_k$.
Functional Tucker~\cite{newman2021functional,ftm_inr} replaces the
discrete row-lookup $U_k[\mathbf{i}]$ with a continuous
coordinate-evaluable basis function $\phi_k:\R\to\R^{r_k}$:
\begin{equation}
u(\rrr;\,G)
  = G\times_1\phi_1(r_1)\times_2\cdots\times_d\phi_d(r_d),
\label{eq:ftm_prelim}
\end{equation}
where $\times_k$ denotes mode-$k$ contraction.
At any fixed coordinate~$\rrr$, the mapping $G\mapsto u(\rrr;\,G)$
is multilinear---linear in each mode separately.
Because $\phi_k$ can be evaluated at arbitrary real-valued inputs, the
decoder supports queries at scattered off-grid sensor locations without
rasterization.
When a single set of basis functions is shared across a field family
and only the core $G$ varies per sample, the model separates common
spatial structure (captured in the bases) from sample-specific
coefficients (captured in the core).

\subsection{Diffusion Priors and Posterior Sampling}
\label{sec:prelim:dps}

A diffusion model places a generative prior $p_\theta(x_0)$ over an
unknown variable $x_0$ by training a denoiser $\epsilon_\theta(x_t,t)$
to reverse a forward noising process
$q(x_t\mid x_0) = \N(\sqrt{\bar\alpha_t}\,x_0,(1{-}\bar\alpha_t)I)$.
Given a noisy state $x_t$, the denoiser produces a clean estimate
$\hat{x}_{0,t} = (x_t - \sqrt{1-\bar\alpha_t}\,\epsilon_\theta(x_t,t))/
\sqrt{\bar\alpha_t}$.
For an inverse problem with partial observations~$\mathbf{y}$ related
to $x_0$ through a forward measurement operator $\mathcal{A}$, the
inference target is the posterior $p(x_0\mid\mathbf{y}) \propto
p_\theta(x_0)\,p(\mathbf{y}\mid x_0)$.
Diffusion Posterior Sampling (DPS)~\cite{chung2023diffusion}
approximates sampling from this posterior by correcting each
unconditional reverse step with a measurement-consistency gradient
evaluated on the clean estimate:
\begin{equation}
x_{t-1} = x'_{t-1} - \zeta_t\,
\nabla_{x_t}\bigl\|\mathbf{y} - \mathcal{A}(\hat{x}_{0,t})\bigr\|_2^2,
\label{eq:dps_posterior}
\end{equation}
where $x'_{t-1}$ is the unconditional reverse sample obtained from
$\hat{x}_{0,t}$ and $\epsilon_\theta(x_t,t)$, and $\zeta_t$ is a
step-size schedule.
The efficiency and accuracy of this guidance depend critically on how
cheaply the measurement operator $\mathcal{A}$ and its gradient can be
evaluated on the clean estimate~$\hat{x}_{0,t}$.


\section{Method}
\label{sec:method}

\begin{figure*}[t]
    \centering
    \includegraphics[width=1.00\textwidth]{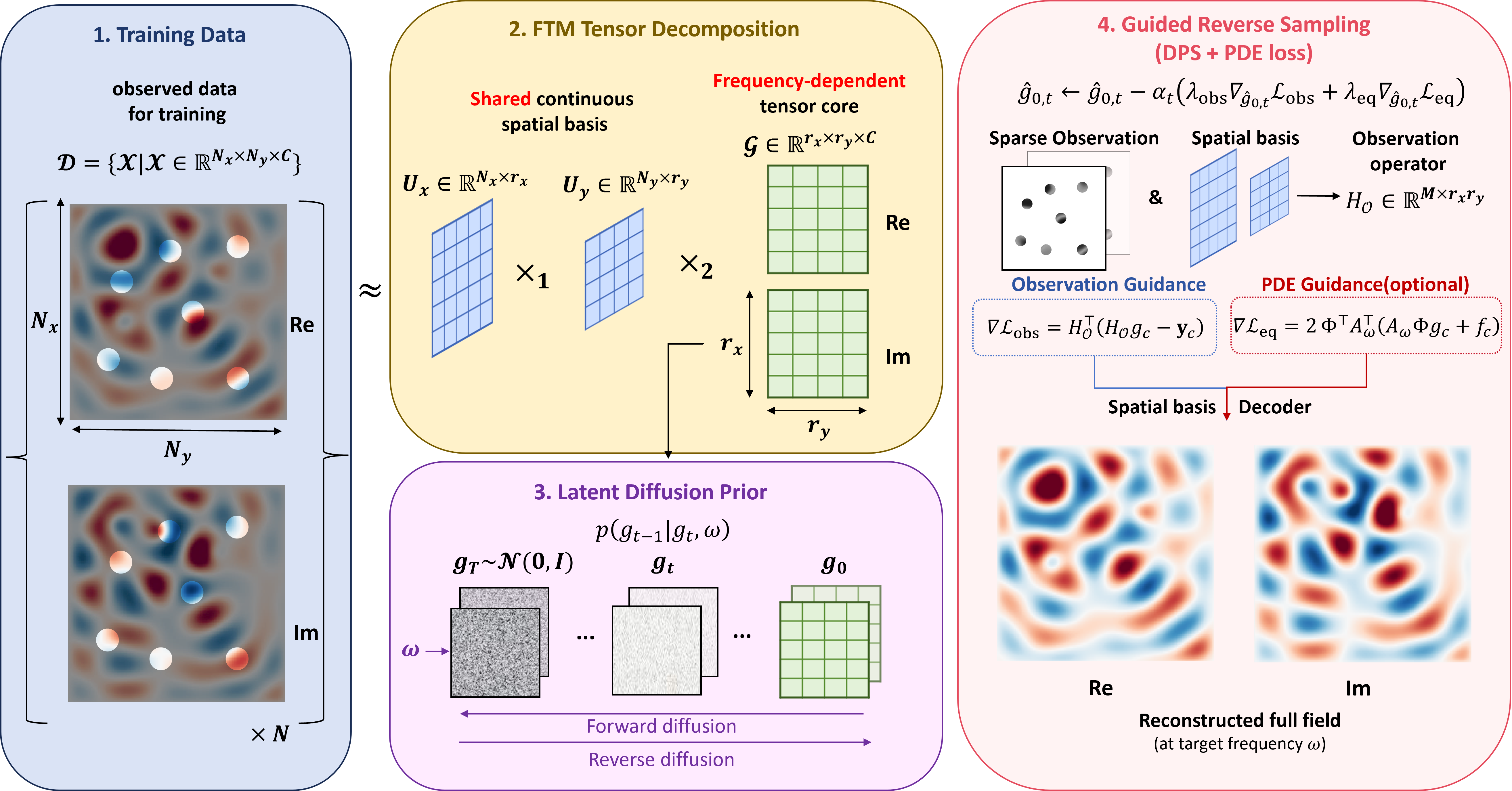}
    \caption{HarmoCore pipeline overview.(2d field as example)
    \emph{Training} (left and center): sparse training observations are
    compressed into per-field joint real--imaginary channel cores via
    shared continuous basis networks;
    normalized cores are used to train a frequency-conditioned
    diffusion prior.
    \emph{Inference} (right): basis evaluation at sensor coordinates
    yields the precomputed observation operator $H_{\mathcal{O}}$;
    core-space posterior sampling guided by $H_{\mathcal{O}}$ and
    an optional equation residual produces the reconstructed complex
    wave field.
    }
    \label{fig:method_overview}
\end{figure*}

HarmoCore reconstructs sparse complex wave fields through two training
stages and a guided inference stage.
\textbf{Training stage 1:} shared continuous spatial basis networks
$(\phi_x,\phi_y)$ and per-field compact cores $\Wcore_{s,\omega}$
are learned jointly from sparse training observations, compressing the field
family into a structured latent.
\textbf{Training stage 2:} a frequency-conditioned diffusion model is
trained on the normalized cores, capturing the distribution of valid
latent coefficients.
\textbf{Inference:} the frozen continuous basis networks are evaluated at the test-case
sensor coordinates, yielding a precomputed observation operator
$H_{\mathcal{O}}$; core-space posterior sampling guided by
$H_{\mathcal{O}}$ and an optional governing-equation residual
reconstructs the core, which is decoded to the continuous wave field.
Figure~\ref{fig:method_overview} illustrates the overall pipeline.

\subsection{Functional Latent Modeling of Complex Wave Fields}
\label{sec:method:rep}

HarmoCore instantiates the Functional Tucker representation of
Eq.~(\ref{eq:ftm_prelim}) with a \emph{joint real--imaginary channel
core}.
For each (sample, frequency) pair $(s,\omega)$, the field is encoded
by a single core tensor and decoded as
\begin{equation}
\begin{aligned}
\Wcore_{s,\omega} &\in \R^{R_x\times R_y\times C},\quad C=2K, \\
u_s(\rrr;\,\Wcore_{s,\omega})
  &= \Wcore_{s,\omega}\times_1\phi_x(r_x)\times_2\phi_y(r_y),
\end{aligned}
\label{eq:harmocore_rep}
\end{equation}
where channels $1,\ldots,K$ are the real parts of the $K$ field
components and channels $K{+}1,\ldots,2K$ are the imaginary parts.
For scalar wave fields $K=1$ ($C=2$).
In 3D a third basis network $\phi_z$ is added and the core lives
in $\R^{R_x\times R_y\times R_z\times C}$.
A single pair of sine-activated MLP networks
$(\phi_x,\phi_y)$~\cite{sitzmann2020siren}
is \emph{shared} across all samples, all frequencies, and all
channels; only $\Wcore_{s,\omega}$ is sample- and frequency-specific.
Each (sample, frequency) pair therefore yields a
$C{\times}R_x{\times}R_y$ core ($R=R_xR_y$ coefficients per channel),
which for the ranks used in our experiments (Section~\ref{sec:exp:setup})
is far smaller than a full-resolution field of size $C{\times}H{\times}W$.

\paragraph{Joint optimization of bases and cores.}
We learn $(\phi_x,\phi_y)$ and all cores simultaneously by Adam
optimization.
Let $\hcore_c = \operatorname{vec}(\Wcore[\cdot,\cdot,c])\in\R^R$
denote the per-channel vectorized core and
$\Phi = [\phi_x(x_i)\otimes\phi_y(y_j)]_{ij}\in\R^{P\times R}$
($P=H{\times}W$) the full-grid basis matrix.
Let $\mathcal{O}$ index the observation coordinates used in the
training objective below; restricting $\Phi$ to the rows indexed by
$\mathcal{O}$ gives the observation operator $H_{\mathcal{O}} =
\Phi[\mathcal{O}] \in\R^{M\times R}$, so that $H_{\mathcal{O}}\hcore_c$
evaluates the decoded field at those positions against the
corresponding measurements $\mathbf{y}_c$.
The training objective is
\begin{equation}
\mathcal{L}_{\mathrm{FTM}}
  = \frac{1}{2}\sum_{c=1}^{C}\mathbb{E}_{s,\omega}\!\left[
    \frac{\|H_{\mathcal{O}}\hcore_c - \mathbf{y}_c\|}
         {\|\mathbf{y}_c\|}\right]
  + \lambda\,\mathcal{S}(\Wcore),
\label{eq:ftm_obj}
\end{equation}
where $\mathcal{O}$ indexes the observed positions and
$\mathcal{S}(\Wcore)$ is a frequency-weighted spatial-smoothness
regularizer on the core matrices.
After training, each core is normalized per channel using global
mean and standard deviation computed over all (sample, frequency)
pairs, yielding zero-mean, unit-variance inputs for the diffusion
model.

\subsection{Frequency-Conditioned Core Diffusion}
\label{sec:method:diffusion}

The normalized cores are used to train a frequency-conditioned
diffusion model.
A conditional UNet $\epsilon_\theta$ takes a noisy core image
$g_t\in\R^{C\times R_x\times R_y}$, diffusion timestep $t$, and
normalized frequency
$\omega_{\mathrm{norm}}=(\omega-\omega_{\min})/(\omega_{\max}-\omega_{\min})
\in[0,1]$ injected via FiLM conditioning at each ResBlock.
The model is trained with the standard noise-prediction
objective~\cite{ho2020ddpm}
\begin{equation}
\begin{aligned}
\mathcal{L}_{\mathrm{diff}}
  &= \mathbb{E}_{g_0,\epsilon,t}\bigl[
    \|\epsilon - \epsilon_\theta(g_t,t,\omega)\|_2^2\bigr], \\
g_t &= \sqrt{\bar\alpha_t}\,g_0 + \sqrt{1-\bar\alpha_t}\,\epsilon,
\end{aligned}
\label{eq:diff_obj}
\end{equation}
using a variance-preserving linear schedule ($T=500$ steps,
$\beta_1=10^{-4}$, $\beta_T=2{\times}10^{-2}$).
This yields the conditional prior $p_\theta(g_0\mid\omega)$ over
normalized joint-channel cores.

\subsection{Core-Space Posterior Reconstruction}
\label{sec:method:dps}

\paragraph{Explicit observation operator.}
At test time, the frozen basis networks are evaluated on the full
evaluation grid to build $\Phi\in\R^{P\times R}$ once, and the
observation operator $H_{\mathcal{O}}=\Phi[\mathcal{O}]$ is
re-instantiated for the test-case sensor coordinates $\mathcal{O}$.
The resulting observation loss and its core-space gradient are
\begin{equation}
\begin{aligned}
\mathcal{L}_{\mathrm{obs}}(\hcore_c)
  &= \|H_{\mathcal{O}}\hcore_c - \mathbf{y}_c\|^2, \\[3pt]
\nabla_{\hcore_c}\mathcal{L}_{\mathrm{obs}}
  &= H_{\mathcal{O}}^\top(H_{\mathcal{O}}\hcore_c - \mathbf{y}_c).
\end{aligned}
\label{eq:hobs_grad}
\end{equation}
Both $H_{\mathcal{O}}$ and the gradient require only matrix--vector
products in $\R^R$, and $H_{\mathcal{O}}$
is reused across all reverse steps.
The same spatial mask applies to all channels (real and imaginary),
so a single $H_{\mathcal{O}}$ serves both.

\paragraph{Core-space sampling update.}
We run a DDPM-style reverse process with guidance applied directly
to the clean estimate at each step:
\begin{equation}
\begin{aligned}
\hat{g}_{0,t}
  &= \frac{g_t - \sqrt{1-\bar\alpha_t}\,\hat\epsilon_\theta}
         {\sqrt{\bar\alpha_t}},
  \quad \hat\epsilon_\theta = \epsilon_\theta(g_t,t,\omega),\\[4pt]
\hat{g}_{0,t}
  &\leftarrow \hat{g}_{0,t}
  - \alpha_t\!\left(\lambda_{\mathrm{obs}}
    \nabla_{\hat{g}_{0,t}}\mathcal{L}_{\mathrm{obs}}
    + \lambda_{\mathrm{eq}}
    \nabla_{\hat{g}_{0,t}}\mathcal{L}_{\mathrm{eq}}\right),\\[4pt]
g_{t-1}
  &= \sqrt{\bar\alpha_{t-1}}\,\hat{g}_{0,t}
   + \sqrt{1-\bar\alpha_{t-1}}\,\hat\epsilon_\theta,
\end{aligned}
\label{eq:dps_update}
\end{equation}
where $\alpha_t$ is a step-specific guidance weight and the noise
estimate $\hat\epsilon_\theta$ is not recomputed after the guidance
correction.

\paragraph{Optional governing-equation guidance.}
For the Helmholtz benchmarks, the discretized operator $A_\omega$
(sparse finite-difference matrix) and source field $f$ are available
from dataset metadata at test time.
The full-grid decoded field per channel is $\hat{u}_c = \Phi\hcore_c$;
the equation residual~\cite{raissi2019pinn} is evaluated on the
decoded dense grid via autograd.
Because $\Phi$ is frozen after FTM training, the decoder Jacobian is
constant across all reverse steps and is reused in each guidance
evaluation without re-evaluation through the basis networks.
For the linear Helmholtz residual $A_\omega\hat{u}+f$, the core-space
gradient is
$\nabla_{\hcore_c}\mathcal{L}_{\mathrm{eq}}
= 2\,\Phi^\top A_\omega^\top(A_\omega\Phi\hcore_c + f_c)$;
in practice we compute this via autograd on the decoded output.
The equation term is optional and supportive:
ablations in Section~\ref{sec:exp:ablation} confirm that removing
observation-guided posterior sampling causes a far larger accuracy
collapse than removing the equation term alone.


\section{Related Work}
\label{sec:related}

Reconstructing dense fields from sparse sensors has been studied
through sensor-to-dense networks, which pool irregular observations
onto a surrogate regular input before prediction~\cite{van2021voronoi},
and neural operators, which learn direct field mappings that
generalize across PDE parameters~\cite{li2021fno,tran2023ffno,lu2021deeponet,li2021pino,li2023geofno,kovachki2023no}.
Both perform well when training coverage captures the relevant
response patterns, but under extreme sparsity ($1$\%--$2$\% sensing)
the oscillatory structure and frequency sensitivity of time-harmonic
wave fields make a single deterministic estimate depend heavily on
whether training instances cover the specific frequency and boundary
configuration.
A generative model that captures the distribution of valid field
configurations can supply the missing structural constraint when local
sensors cannot resolve the global wave pattern alone.

Diffusion models~\cite{ho2020ddpm} supply such priors and can be
combined with partial observations through posterior
sampling~\cite{song2021score,chung2023diffusion,kawar2022ddrm,song2022medicalscore,song2023piGDM}.
building on efficient and controllable sampling
techniques~\cite{song2021ddim,ho2022classifierfree};
Diffusion Posterior Sampling~\cite{chung2023diffusion} steers a
learned denoiser toward measurement-consistent states during the
reverse process, and related work applies this directly in pixel space
to PDE-governed field completion~\cite{gao2024diffusionpde}.
Modeling and guiding every value of the dense spatial field is costly
for rapidly varying, channel-coupled complex fields, and grows more
expensive as resolution or dimensionality increases---motivating a
compact, continuous representation aligned with the real/imaginary
channel coupling, in the spirit of latent-space
diffusion~\cite{rombach2022ldm}.

Tucker and tensor-train methods offer compact multi-mode decompositions
of structured fields~\cite{kolda2009tensor,oseledets2011tensortrain,newman2021functional},
and Functional Tucker models~\cite{ftm_inr} replace discrete factor
matrices with continuous-coordinate basis
functions~\cite{tancik2020fourier,mildenhall2020nerf}, separating
shared spatial variation (the bases) from sample-specific coefficients
(the core).
Closest to our work, \cite{sdift2025} combines learned Functional
Tucker cores with a diffusion prior for spatiotemporal field
reconstruction.
HarmoCore instead targets time-harmonic complex wave fields under
extreme-sparse scattered sensing, conditioning the diffusion prior on
frequency, using a joint real--imaginary channel core, and exploiting
the multilinear decoder at fixed sensor coordinates to form an explicit
observation operator for efficient core-space posterior
sampling---keeping guidance in a compact, structured latent rather than
the dense pixel space.


\section{Experiments}
\label{sec:experiments}

\begin{figure*}[!tb]
    \centering
    \includegraphics[width=1.00\textwidth]{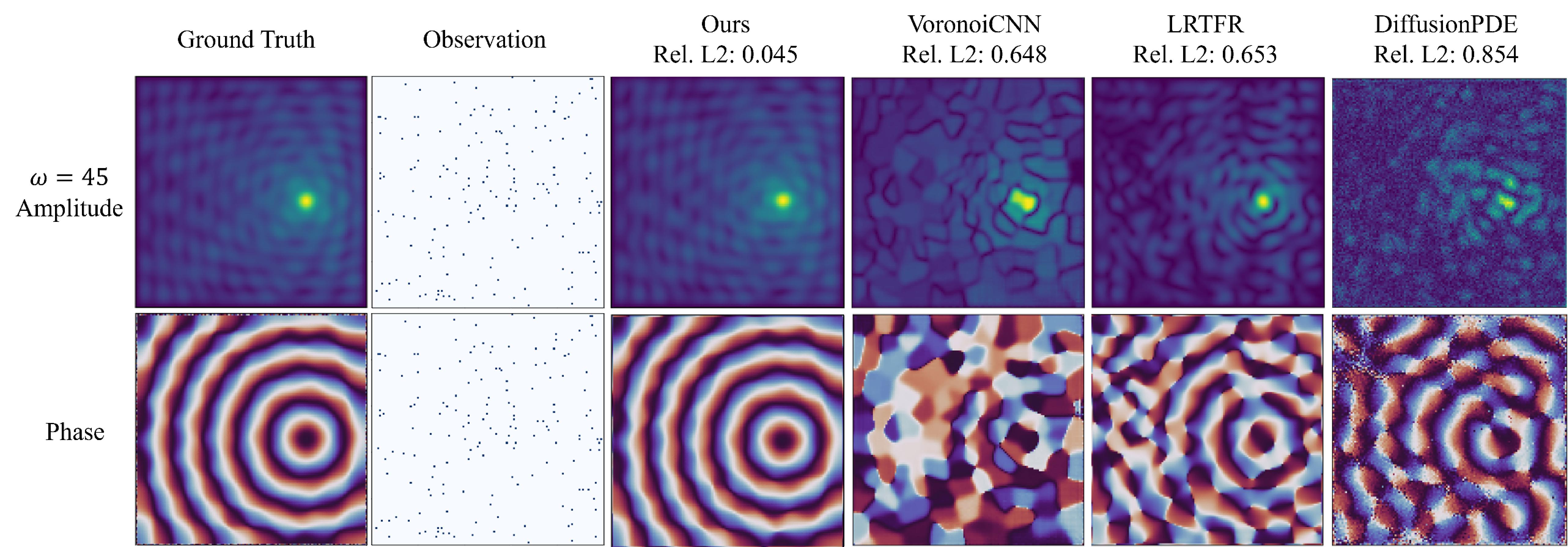}
    \caption{Qualitative comparison on 2D Helmholtz at extreme sparsity
    ($1\%$ sensing).
    HarmoCore recovers the global oscillatory structure more faithfully
    than competing baselines at this sensing density.
    }
    \label{fig:helmholtz2d_qual}
\end{figure*}

\subsection{Experimental Setup}
\label{sec:exp:setup}

We evaluate on three benchmarks, each isolating a different
experimental question.
\textbf{2D Helmholtz} is the primary benchmark and drives the main
sparsity comparison across sensor densities.
\textbf{2D Synthetic Wave Fields} test whether the same core-space
formulation transfers to a physically different wave-field family
generated from a closed-form ray model rather than a PDE solve: each
field is a direct-plus-reflected superposition of per-source ray terms,
\begin{equation}
    \begin{aligned}
\tilde u_s(\rrr,\omega) = \sum_{k=1}^{K_s} w_{s,k}\Bigl[
a_1(\rrr,\rrr_{s,k})\,e^{\,i2\pi f\,\tau_1(\rrr,\rrr_{s,k})}
\\
+ \beta\, a_2(\rrr,\rrr_{s,k}')\,e^{\,i2\pi f\,\tau_2(\rrr,\rrr_{s,k}')}
\Bigr],
\end{aligned}
\label{eq:ray_field}
\end{equation}
with $\omega=2\pi f$, $a_1,a_2$ distance-dependent amplitude decays,
$\tau_1,\tau_2$ the direct/reflected travel times, and
$\rrr_{s,k}'$ each source's mirror-image reflector (full
parameterization in Appendix~\ref{app:data}); having no governing
equation, this benchmark receives no physics-residual metric or
equation guidance (Section~\ref{sec:method:dps}).
\textbf{3D Helmholtz} reuses the same governing equation on a
higher-dimensional domain to test whether the framework extends to 3D via the natural addition of a third 
shared basis network $\phi_z$.
All methods on a given benchmark are evaluated on the same held-out
test set.

At each sensing ratio $r\in\{1\%,2\%,5\%,10\%\}$, the observation mask
is an i.i.d.\ Bernoulli($r$) mask over grid points---each grid location
is retained as an observed sensor independently with probability
$r$---generated once per ratio and shared by every method on a given
test case.
We report $1\%$, $2\%$, and $5\%$, the extreme-to-moderate
sparsity range that is the focus of our method.

We compare against three categories of baselines: a continuous
low-rank tensor-function representation, LRTFR~\cite{luo2024lrtfr};
deterministic operator-regression networks trained on dense
observations, FNO~\cite{li2021fno}, F-FNO~\cite{tran2023ffno}, and
VoronoiCNN~\cite{van2021voronoi}; and a generative diffusion-based
baseline, DiffusionPDE~\cite{gao2024diffusionpde}.

The primary metric is the relative reconstruction error (relative
$\ell_2$ norm, mean$\pm$std over the test set), which we refer to
throughout as \emph{Relative L2 Error} (\emph{Rel.\ L2} in tables); it
is computed over the \emph{full} evaluation grid including sensor
locations, and the primary comparison focuses on $1\%$--$2\%$ sensing,
where the inverse problem is most underdetermined.
For the two Helmholtz benchmarks we additionally report a
physics-residual metric under the discretized Helmholtz operator; the
2D Synthetic benchmark has no governing PDE, so this column is marked
``---'' for all methods there.
Lower values are better for all metrics.
Unless otherwise noted, the Functional Tucker core uses ranks
$R_x=R_y=24$ ($R_x=R_y=R_z=24$ in 3D), yielding $R=576$ (2D) or
$R=13{,}824$ (3D) coefficients per channel per (sample, frequency)
pair (Section~\ref{sec:method:rep}). Details of the experimental setup
is given in Appendix~\ref{app:impl}.

\begin{table*}[t]
\centering
\caption{Reconstruction Relative L2 Error and physics residual across
three benchmarks (mean$\pm$std).
Physics residual is a PDE residual under the Helmholtz operator for
2D/3D Helmholtz; the 2D Synthetic benchmark has no governing PDE
(fields are generated by a closed-form ray model), so this column is
``---'' for all methods there.
}
\label{tab:main_all}
\setlength{\tabcolsep}{4pt}
\begin{tabular}{lcccccc}
\toprule
& \multicolumn{2}{c}{1\%} & \multicolumn{2}{c}{2\%} & \multicolumn{2}{c}{5\%} \\
\cmidrule(lr){2-3} \cmidrule(lr){4-5} \cmidrule(lr){6-7}
Method & Rel.\ L2 & Phys. Res. & Rel.\ L2 & Phys. Res. & Rel.\ L2 & Phys. Res. \\
\midrule
\multicolumn{7}{l}{\textit{2D Helmholtz}} \\
\textbf{Ours} & \textbf{0.130$\pm$0.069} & \textbf{0.032$\pm$0.001} & \textbf{0.068$\pm$0.039} & \textbf{0.032$\pm$0.001} & \textbf{0.037$\pm$0.023} & \textbf{0.032$\pm$0.001} \\
LRTFR & 0.726$\pm$0.104 & 0.181$\pm$0.164 & 0.505$\pm$0.142 & 0.172$\pm$0.168 & 0.251$\pm$0.120 & 0.138$\pm$0.119 \\
FNO & 0.822$\pm$0.099 & 1.518$\pm$0.906 & 0.609$\pm$0.060 & 1.494$\pm$0.916 & 0.239$\pm$0.031 & 0.930$\pm$0.599 \\
F-FNO & 1.413$\pm$0.403 & 1.338$\pm$0.556 & 1.413$\pm$0.407 & 1.841$\pm$0.760 & 1.412$\pm$0.418 & 2.869$\pm$1.192 \\
VoronoiCNN & 0.638$\pm$0.298 & 0.201$\pm$0.048 & 0.446$\pm$0.227 & 0.186$\pm$0.038 & 0.222$\pm$0.123 & 0.153$\pm$0.024 \\
DiffusionPDE & 0.653$\pm$0.203 & 1.114$\pm$0.326 & 0.438$\pm$0.197 & 1.013$\pm$0.279 & 0.179$\pm$0.118 & 0.745$\pm$0.211 \\
\midrule
\multicolumn{7}{l}{\textit{2D Synthetic Wave Fields}} \\
\textbf{Ours} & \textbf{0.259$\pm$0.106} & --- & \textbf{0.102$\pm$0.049} & --- & \textbf{0.035$\pm$0.013} & --- \\
LRTFR & 0.503$\pm$0.038 & --- & 0.224$\pm$0.017 & --- & 0.170$\pm$0.070 & --- \\
FNO & 1.101$\pm$0.034 & --- & 1.091$\pm$0.033 & --- & 1.058$\pm$0.032 & --- \\
F-FNO & 1.075$\pm$0.049 & --- & 1.070$\pm$0.050 & --- & 1.056$\pm$0.051 & --- \\
VoronoiCNN & 0.440$\pm$0.200 & --- & 0.278$\pm$0.136 & --- & 0.122$\pm$0.061 & --- \\
DiffusionPDE & 0.434$\pm$0.169 & --- & 0.194$\pm$0.120 & --- & 0.038$\pm$0.021 & --- \\
\midrule
\multicolumn{7}{l}{\textit{3D Helmholtz}} \\
\textbf{Ours} & \textbf{0.249$\pm$0.208} & \textbf{5.85$\pm$3.69} & \textbf{0.175$\pm$0.140} & \textbf{5.34$\pm$3.15} & \textbf{0.141$\pm$0.112} & \textbf{5.12$\pm$2.96} \\
LRTFR & 3.139$\pm$2.293 & 603$\pm$1604 & 2.494$\pm$5.407 & 816$\pm$3032 & 5.575$\pm$3.221 & 710$\pm$1333 \\
FNO & 1.217$\pm$0.125 & 127$\pm$51 & 1.212$\pm$0.125 & 245$\pm$105 & 1.197$\pm$0.123 & 576$\pm$252 \\
F-FNO & 1.144$\pm$0.095 & 117$\pm$42 & 1.144$\pm$0.098 & 226$\pm$83 & 1.146$\pm$0.106 & 523$\pm$196 \\
VoronoiCNN & 0.440$\pm$0.162 & 14.8$\pm$8.0 & 0.291$\pm$0.109 & 11.7$\pm$6.5 & 0.161$\pm$0.060 & 8.47$\pm$4.92 \\
DiffusionPDE & 0.753$\pm$0.192 & 84.9$\pm$37.0 & 0.511$\pm$0.154 & 81.7$\pm$42.3 & 0.251$\pm$0.069 & 67.6$\pm$37.2 \\
\bottomrule
\end{tabular}
\end{table*}

\subsection{Main Results across Benchmarks}
\label{sec:exp:main}
  Table~\ref{tab:main_all} reports reconstruction Rel.\ L2 error and
  physics residual across all three benchmarks.
  HarmoCore achieves the lowest Rel.\ L2 error and best physics consistency on every benchmark at
  all three sensing ratios ($1\%$, $2\%$, and $5\%$), with the largest
  margin over the best baseline at $1\%$--$2\%$ sensing, where the
  inverse problem is least constrained. Figure~\ref{fig:helmholtz2d_qual} and Figure~\ref{fig:helmholtz3d_qual_app} show representative qualitative examples. Full qualitative examples are shown in Appendix~\ref{app:qual_more}.

  \begin{figure}[t]
      \centering
      \includegraphics[width=1.00\linewidth]{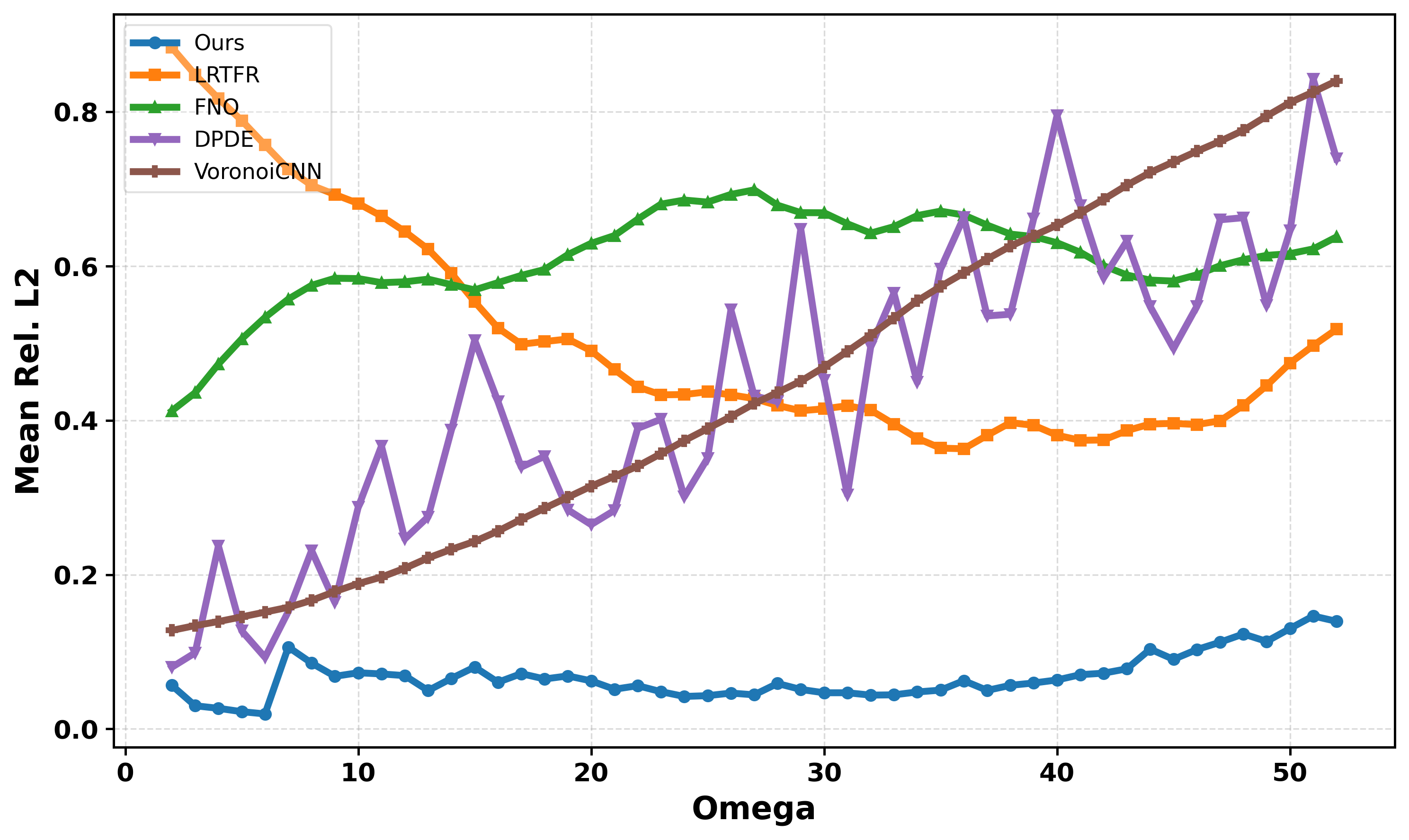}
      \caption{2D Helmholtz mean reconstruction Relative L2 Error vs.\
      frequency $\omega$ at $2\%$ observation rate. HarmoCore (Ours) stays
      low and comparatively flat across the frequency range, while
      baseline error fluctuates irregularly with $\omega$.}
      \label{fig:freq_curve_2pct}
  \end{figure}
  Figure~\ref{fig:freq_curve_2pct} further breaks this down by frequency
  at $2\%$ sensing: HarmoCore's error stays low and comparatively flat
  across the tested frequency range, while the baselines fluctuate
  irregularly with $\omega$ rather than following a stable trend,
  underscoring HarmoCore's comparative robustness across the frequency
  range.

  \begin{figure*}[t]
      \centering
      \includegraphics[width=1.00\textwidth]{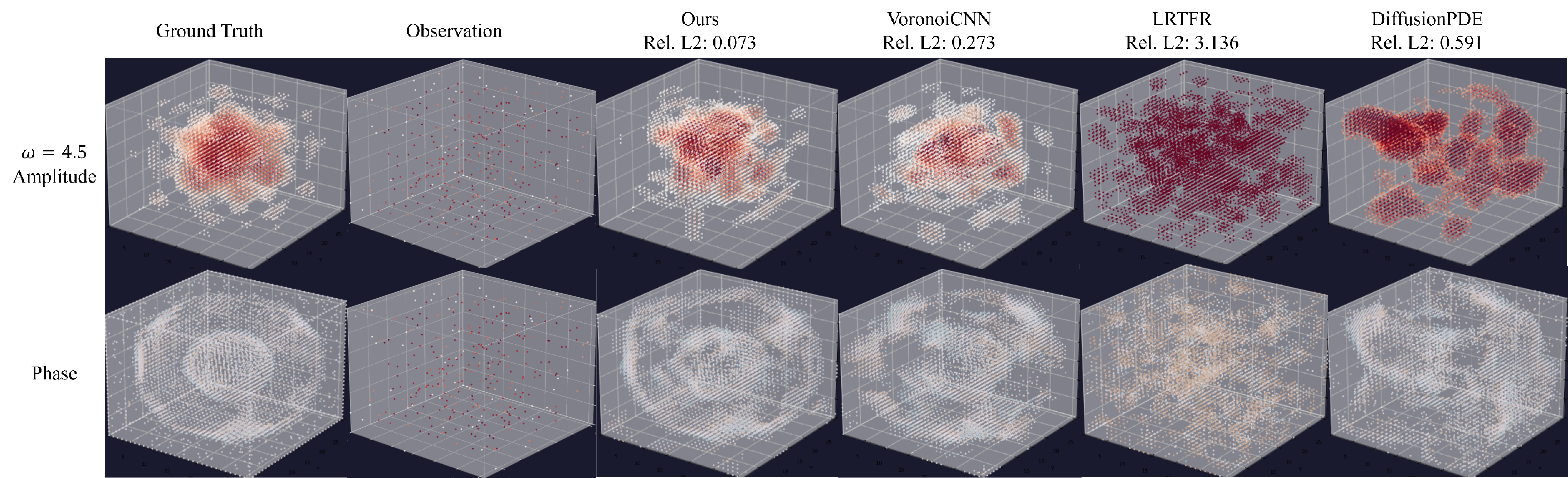}
      \caption{Qualitative 3D Helmholtz reconstructions at $1\%$ sensing
      , comparing ground truth, HarmoCore, and the
      strongest baselines.
      HarmoCore recovers the correct interference and phase structure
      throughout the volume, while baselines flatten detail or drift in
      phase away from observed sensors.
      }
      \label{fig:helmholtz3d_qual_app}
  \end{figure*}

  A consistent pattern across all three benchmarks is that HarmoCore is
  most valuable in the truly underdetermined regime: when the sensor
  ratio is extremely low, sparse observations alone do not constrain the
  full field, and the learned core-space prior resolves this ambiguity,
  which is exactly where HarmoCore's margin over every baseline is
  largest.

\subsection{Mechanism Analysis and Ablation}
\label{sec:exp:ablation}

Table~\ref{tab:ablation_helmholtz} reports ablations on 2D Helmholtz.
Removing DPS guidance entirely---keeping only the PDE-residual
term---causes a large accuracy collapse at both $1\%$ and $2\%$,
confirming that the learned prior and observation-guided posterior
sampling are the primary source of reconstruction accuracy.
Removing only the equation term produces a smaller but notable
degradation, particularly at $1\%$ sensing, which supports the
interpretation that equation guidance acts as a supplementary physical
regularizer rather than the primary reconstruction mechanism.

The ``w/o~DPS'' variant achieves a lower PDE residual than the full
method despite substantially worse reconstruction accuracy.
This is expected: that variant optimizes directly toward equation
consistency while receiving no constraint from observations.
A low PDE residual alone is not sufficient for correct field recovery
under extreme sparsity; physical consistency must be interpreted
jointly with reconstruction error.

\begin{table}[t]
\centering
\caption{Ablation on 2D Helmholtz (lower is better).
}
\label{tab:ablation_helmholtz}
\begin{tabular}{lcccc}
\toprule
& \multicolumn{2}{c}{1\%} & \multicolumn{2}{c}{2\%} \\
\cmidrule(lr){2-3}\cmidrule(lr){4-5}
Method & Rel.\ L2 & Phys. Res. & Rel.\ L2 & Phys. Res. \\
\midrule
\textbf{Ours} & \textbf{0.130} & 0.032 & \textbf{0.068} & 0.032 \\
w/o DPS & 0.960 & \textbf{0.027} & 0.921 & \textbf{0.028} \\
w/o PDE & 0.431 & 0.058 & 0.204 & 0.044 \\
\bottomrule
\end{tabular}
\end{table}

Together, these ablations indicate that the dominant ingredient is the
combination of a compact Functional Tucker core with observation-guided
DPS: the shared continuous basis reduces the dimensionality of the
inverse problem while the diffusion prior and DPS inject the actual
observations into the posterior reconstruction, whereas the
PDE-residual term acts mainly as a supporting constraint that improves
physical consistency without being the main reason the method works.

\subsection{Additional Analyses}
\label{sec:exp:additional}

\paragraph{Representation capacity.}
\begin{figure}[t]
    \centering
    \includegraphics[width=1.00\linewidth]{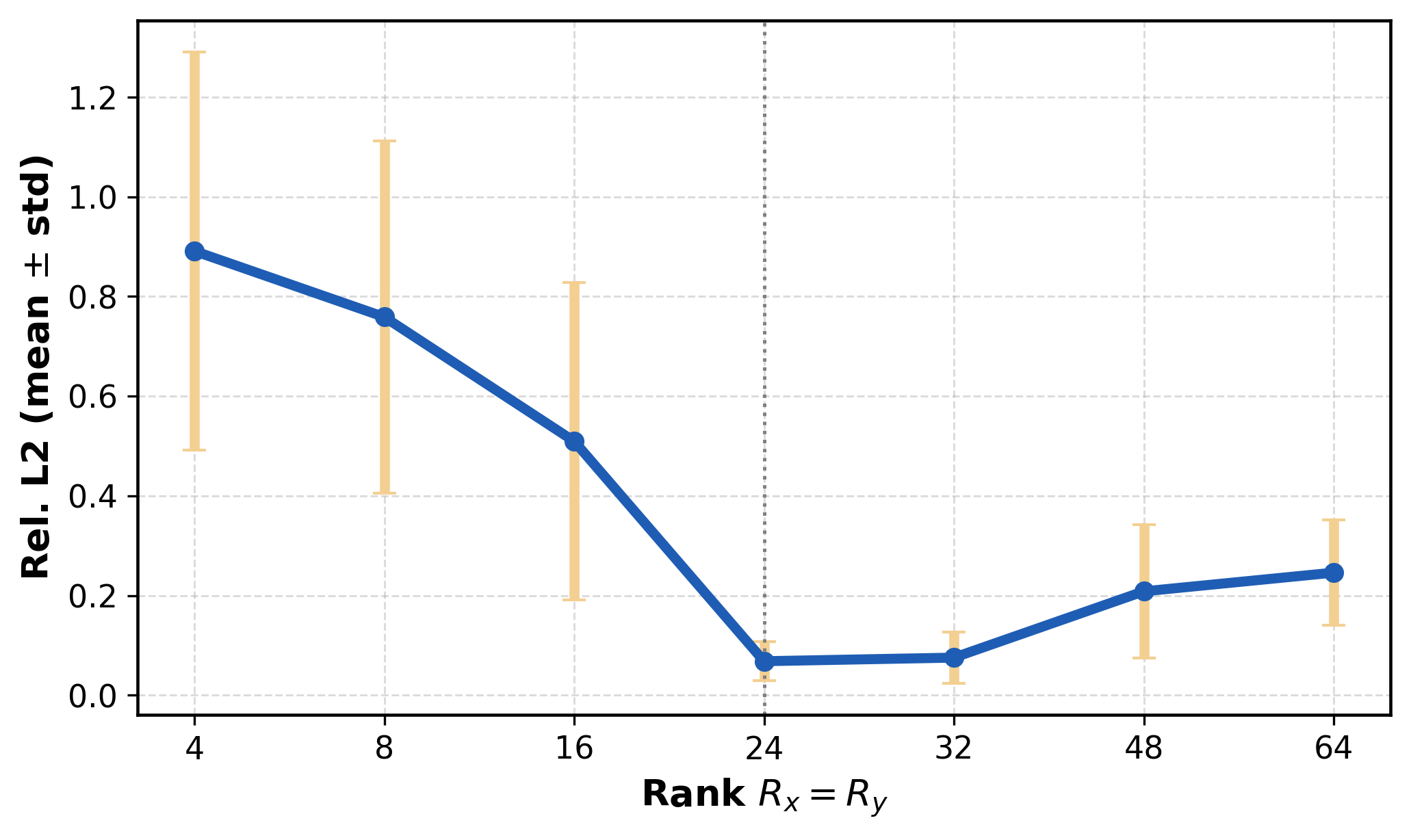}
    \caption{Reconstruction Relative L2 Error (mean$\pm$std) vs.\
    Functional Tucker rank $R_x=R_y$ on 2D Helmholtz at $2\%$ sensing
    ($102$ test cases per rank), with the diffusion-prior architecture
    and training configuration held fixed across ranks. The default
    rank $R=24$ (dashed line) attains the lowest error under this fixed
    training budget.}
    \label{fig:rank_sweep}
\end{figure}
Figure~\ref{fig:rank_sweep} sweeps the Functional Tucker rank
$R_x=R_y\in\{4,8,16,24,32,48,64\}$ on 2D Helmholtz at $2\%$ sensing,
retraining the diffusion prior at each rank with the same architecture
and training budget. Reconstruction error is non-monotonic in rank: it
falls sharply from $R{=}4$ ($0.891\pm0.400$) to $R{=}16$
($0.509\pm0.319$), reaches its minimum at the default $R{=}24$
($0.068\pm0.039$, matching Table~\ref{tab:main_all}), stays close at
$R{=}32$ ($0.075\pm0.052$), and rises again at $R{=}64$
($0.246\pm0.106$). Since a larger core is at least as expressive as a
smaller one, this reflects a fixed-budget diffusion prior becoming
increasingly under-provisioned as the core grows, rather than an
intrinsic capacity ceiling---we read $R{=}24$ as the best operating
point under the current, rank-independent training budget.

\paragraph{Robustness to distribution shift.}
To test robustness to a shift in the underlying generative distribution
at test time, we evaluate all methods, without retraining, on an
out-of-distribution (OOD) variant of the 2D Synthetic benchmark that
redraws the per-sample wave speed from a substantially wider range
(full construction in Appendix~\ref{app:ood_synth}).
Table~\ref{tab:ood_synth} reports Rel.\ L2 error under this shift:
HarmoCore's error is essentially unchanged from its in-distribution
values and LRTFR is similarly robust, while the remaining baselines
degrade substantially, most sharply DiffusionPDE.

\begin{table}[t]
\centering
\caption{Reconstruction Rel.\ L2 error on the 2D Synthetic OOD test
set (shifted wave-speed range, no retraining).}
\label{tab:ood_synth}
\begin{tabular}{lccc}
\toprule
Method & 1\% & 2\% & 5\% \\
\midrule
\textbf{Ours} & \textbf{0.254} & \textbf{0.097} & \textbf{0.033} \\
LRTFR & 0.496 & 0.206 & 0.174 \\
FNO & 1.119 & 1.110 & 1.083 \\
F-FNO & 1.087 & 1.083 & 1.069 \\
VoronoiCNN & 0.455 & 0.307 & 0.154 \\
DiffusionPDE & 0.542 & 0.326 & 0.140 \\
\bottomrule
\end{tabular}
\end{table}


\section{Conclusion}
\label{sec:conclusion}

We presented HarmoCore, a latent generative framework for sparse
wave-field reconstruction representing complex fields as a compact
Functional Tucker core over shared continuous spatial bases and
performs diffusion posterior sampling directly in this core space
rather than dense pixel space.
Across 2D Helmholtz, 2D synthetic wave fields, and 3D Helmholtz,
this formulation is most effective in the most underdetermined
regimes, where deterministic reconstruction and prior-free low-rank
fitting fail, while maintaining better physical
consistency than dense operator and pixel-space generative baselines.
The method relies on globally learned spatial
parameterization and on first obtaining a sufficiently expressive FTM
representation, and the paper offers stronger evidence for sparse
reconstruction than for frequency extrapolation or uncertainty
calibration; extending the core-space prior along these directions is
left to future work.

\bibliography{references}
\clearpage

\appendix

\section{Implementation Details}
\label{app:impl}

This appendix collects implementation details that are omitted from the
main paper for space reasons.

\subsection{Datasets and Preprocessing}
\label{app:data}

All three benchmarks (2D Helmholtz, 2D synthetic wave fields, 3D
Helmholtz; Section~\ref{sec:exp:setup}) use scalar complex fields
($K=1$, $C=2$: one real and one imaginary channel per
(sample, frequency) pair; Eq.~\eqref{eq:field_obs}).

All methods in Table~\ref{tab:main_all} are evaluated on the same
held-out test set per benchmark: $153$ held-out fields for 2D
Helmholtz, $170$ held-out cases ($10$ samples $\times$ $17$
frequencies) for 2D Synthetic, and $410$ held-out cases ($10$ samples
$\times$ $41$ frequencies) for 3D Helmholtz.

Sensor ratios of $1\%$, $2\%$, and $5\%$ of the evaluation grid are
used in the main paper; the $10\%$ ratio is reported in
Appendix~\ref{app:full_tables}.

After FTM training, each core is normalized per channel using the
global mean and standard deviation computed over all
(sample, frequency) pairs (Section~\ref{sec:method:rep}), producing
zero-mean, unit-variance inputs for diffusion training.

\paragraph{2D and 3D Helmholtz generation.}
Both Helmholtz benchmarks numerically solve the PML-damped Helmholtz
equation
\begin{equation}
\Delta\tilde u_s(\rrr,\omega) + (\omega/c)^2\,\tilde u_s(\rrr,\omega)
= -f_s(\rrr,\omega),
\label{eq:helmholtz_pde}
\end{equation}
with $L=1$, $c=1$, $\rrr\in\spdom = [0,L]^d$, $d\in\{2,3\}$, discretized by second-order finite differences on a
uniform $128^2$ grid (2D) or $32^3$ grid (3D).
$\Delta$ is realized as a complex-coordinate-stretched Laplacian
$\sum_{k=1}^d \partial_{r_k}[\,s_k(r_k)\,\partial_{r_k}\,]$ that
implements a quadratic-profile PML absorbing layer of width $\eta L$
on every face ($\eta=0.12$ in 2D, $0.15$ in 3D):
\begin{equation}
\sigma(r_k) = \sigma_{\max}
\Bigl(\tfrac{\eta L - \operatorname{dist}(r_k,\partial\spdom)}{\eta L}\Bigr)^{2}
,
\label{eq:pml_stretch}
\end{equation}
with $\sigma_{\max}=50$ (2D) or $40$ (3D); $\tilde u_s=0$ is imposed on
the outermost grid rows (Dirichlet).
The source term is a sum of $K_s$ random Gaussian point sources with
unit-magnitude, random-phase complex amplitudes,
\begin{equation}
f_s(\rrr,\omega) = \sum_{k=1}^{K_s} e^{i\phi_{s,k}}\,
\exp\!\bigl(-\|\rrr-\rrr_{s,k}\|^2/2\sigma_{\mathrm{src}}^2\bigr),\\ 
\label{eq:gaussian_source}
\end{equation}
with $\phi_{s,k}\sim\mathcal{U}(0,2\pi)$, $K_s\in\{1,\dots,4\}$ (2D, $\sigma_{\mathrm{src}}=0.025$) or
$K_s\in\{1,2,3\}$ (3D), source positions $\rrr_{s,k}$ drawn uniformly
inside $\spdom$ away from the PML layer, and held fixed across every
frequency of sample $s$ (only $\omega$ changes the linear system, so
$(\rrr_{s,1{:}K_s},\phi_{s,1{:}K_s})$ is the per-sample instance
descriptor $\xi_s$ in $\mathcal{L}_{\xi_s,\omega}$, Section~\ref{sec:prelim:task}).
Frequencies are sampled on a linear grid: 2D Helmholtz uses
$\omega\in[2,52]$ and 3D Helmholtz uses $\omega\in[2,22]$.
Each complex solution is split into real/imaginary channels and the
whole dataset is divided by a single global scale (the maximum
absolute value over all samples, frequencies, and grid points) before
FTM fitting.

\paragraph{2D synthetic wave-field generation.}
The synthetic benchmark replaces the PDE solve with a closed-form
direct-plus-reflected ray superposition on the same $128\times128$
grid over $\spdom=[0,1]^2$.
Sample $s$ draws a fixed wave speed $v_s\sim\mathcal{U}(0.8,1.2)$ and
$K_s\in\{1,2,3\}$ source positions $\rrr_{s,k}$ (uniform in $\spdom$),
each held fixed across that sample's frequency sweep; every source has
a mirror-image reflector $\rrr_{s,k}'$ across the boundary $r_2=0$ and
a random weight $w_{s,k}\sim\mathcal{U}(0.8,1.2)$.
The field at frequency $\omega=2\pi f$ is given by
Eq.~\eqref{eq:ray_field} (Section~\ref{sec:exp:setup}), where
$\beta=0.18$ scales the reflected term; the direct/reflected
travel times are
$\tau_1 = (r_1/v_s)\bigl(1+\varepsilon\,\Phi_1(\rrr)\bigr)$ and
$\tau_2 = \bigl((r_2+\delta)/v_s\bigr)\bigl(1+\varepsilon\,\Phi_2(\rrr)\bigr)$,
with $r_1,r_2$ the distances from $\rrr$ to the source and its
reflector, $\delta=0.35$ a fixed delay bias, $\varepsilon=0.08$ a
phase-perturbation strength, and $\Phi_1,\Phi_2$ fixed smooth
sinusoidal fields (in $\sin,\cos$ of $r_1,r_2$ over the domain extent)
shared by every source in a sample; the amplitude decays are
$a_1 = e^{-\alpha_1 r_1}/(r_1+r_0)^{p}$ and
$a_2 = e^{-\alpha_2 r_2}/\sqrt{r_2}$, with $\alpha_1=0$, $\alpha_2=0.12$,
$r_0=0.4$, $p=0.3$.
In the unperturbed limit ($\varepsilon\to0$) the direct-wave phase
satisfies the eikonal relation $\|\nabla\theta\|=2\pi f/v_s$; because
the field is produced by this closed-form summation rather than a
governing-equation solve, no PDE residual is available for this
benchmark (Section~\ref{sec:method:dps}).
The evaluation frequency grid is $17$ linearly spaced points in
$[1,5]\,\mathrm{Hz}$; data are globally rescaled by the maximum
absolute value, as in the Helmholtz benchmarks.

\subsection{Functional Tucker Representation}
\label{app:ftm_impl}

The shared spatial bases $(\phi_x,\phi_y)$ (and $\phi_z$ in 3D) are
sine-activated SIREN MLPs~\cite{sitzmann2020siren}, one pair (triple
in 3D) shared across all samples, frequencies, and channels
(Section~\ref{sec:method:rep}).
Default ranks are $R_x=R_y=24$, giving $R=R_xR_y=576$ coefficients per
channel per (sample, frequency) pair, versus a full-resolution field
of size $C\times H\times W$ (Eq.~\eqref{eq:harmocore_rep}).
Bases and all per-(sample, frequency) cores are optimized jointly with
Adam, using the objective in Eq.~\eqref{eq:ftm_obj}: a relative
reconstruction loss on the observed positions plus a
frequency-weighted spatial-smoothness regularizer $\mathcal{S}(\Wcore)$
(a spatial-gradient penalty weighted by normalized frequency).
Both the basis networks and the cores use learning rate $10^{-4}$,
with a batch size of $64$ samples over $25{,}000$ training iterations;
the smoothness regularizer weight is $\lambda=10^{5}$ in
Eq.~\eqref{eq:ftm_obj}.
Each SIREN basis network has $4$ hidden layers of width $512$.


\subsection{Latent Diffusion Model}
\label{app:diff_impl}

The diffusion prior is a conditional UNet $\epsilon_\theta$ operating
on the normalized joint-channel core image
$g_t\in\R^{C\times R_x\times R_y}$ (Section~\ref{sec:method:diffusion}).
Frequency conditioning uses the normalized scalar
$\omega_{\mathrm{norm}}\in[0,1]$ injected via FiLM at each ResBlock.
Training uses the noise-prediction objective (Eq.~\eqref{eq:diff_obj})
under a variance-preserving linear schedule with $T=500$ steps,
$\beta_1=10^{-4}$, $\beta_T=2\times10^{-2}$.
Optimization uses AdamW with learning rate $10^{-4}$, weight decay
$10^{-6}$, and a batch size of $32$, trained for $500$ epochs; no EMA
of model weights is used.
We additionally tried richer frequency encodings (Fourier features)
during development; these did not yield consistent gains over the
scalar-FiLM default and are treated as a negative result
(Appendix~\ref{app:freq_conditioning}).
\subsection{Posterior Sampling and Guidance}
\label{app:sampling_impl}

At inference, the frozen bases are evaluated once at the test-case
sensor coordinates to build the observation operator $H_{\mathcal{O}}$
(Eq.~\eqref{eq:hobs_grad}), which is reused across all reverse steps.
Guidance is applied directly to the clean estimate $\hat{g}_{0,t}$ at
each reverse step (Eq.~\eqref{eq:dps_update}), combining the
observation-likelihood gradient (weight $\lambda_{\mathrm{obs}}$) and,
optionally, a governing-equation residual gradient (weight
$\lambda_{\mathrm{eq}}$); the step-specific scale $\alpha_t$ multiplies
the combined correction.
Observations are treated as noiseless ($\sigma_{\mathrm{obs}}=0$) in
all experiments, so $\lambda_{\mathrm{obs}}$ absorbs the likelihood
scaling.
For the Helmholtz benchmarks, the equation term uses the closed-form
residual gradient given in Section~\ref{sec:method:dps}. The 2D
Synthetic benchmark has no governing PDE, so no equation-guidance term
is applied there ($\lambda_{\mathrm{eq}}=0$); reconstruction on that
benchmark uses the observation term only.

\subsection{Evaluation Metrics}
\label{app:metrics}

For a predicted field $\hat u$ and ground truth $u$ (channel-stacked
real/imaginary parts, size $C\times H\times W[\times D]$), the
reconstruction error reported throughout the main tables is the
relative $\ell_2$ error over the full evaluation grid, which we refer
to as \emph{Relative L2 Error} (\emph{Rel.\ L2}),
\begin{equation}
\mathrm{Rel.\,L2} = \frac{\|\hat u - u\|_2}{\|u\|_2},
\label{eq:rmse_def}
\end{equation}
computed jointly over all channels and grid points---including sensor
locations---then aggregated as mean$\pm$std over the test set; this
full-field definition is used consistently for HarmoCore and every
baseline.

For the two Helmholtz benchmarks, the physics-residual metric
evaluates the discretized governing operator $A_\omega$
(Eq.~\eqref{eq:helmholtz_pde}) against the predicted field and known
source $f$ on interior grid points $\mathcal{I}$ (the domain with the
outermost boundary row/column excluded):
\begin{equation}
\mathrm{PhysRes} = \sqrt{\frac{\operatorname{mean}_{\mathcal{I}}
\bigl(|A_\omega\hat u + f|^2\bigr)}
{\operatorname{mean}_{\mathcal{I}}\bigl(|f|^2\bigr)}}.
\label{eq:physres_def}
\end{equation}

\subsection{Baseline Settings}
\label{app:baseline_settings}

All learned baselines are trained on the same $80\%/20\%$
train/validation split of the dense training set for each benchmark
(Appendix~\ref{app:data}) and evaluated on the same held-out test set
as HarmoCore.

\textbf{FNO}~\cite{li2021fno}: $4$ Fourier layers, width $64$, $12$
Fourier modes per spatial dimension, trained for $200$ epochs (Adam,
learning rate $10^{-3}$, batch size $64$).

\textbf{F-FNO}~\cite{tran2023ffno}: $4$ factorized Fourier layers,
width $64$, $12$ modes per spatial dimension, trained for $200$ epochs
(Adam, learning rate $10^{-3}$, batch size $64$).

\textbf{VoronoiCNN}~\cite{van2021voronoi}: convolutional
encoder--decoder, base width $64$, trained for $200$ epochs (Adam,
learning rate $10^{-3}$, batch size $32$).

\textbf{DiffusionPDE}~\cite{gao2024diffusionpde}: conditional UNet
(base width $64$, conditioning dimension $256$) trained with the same
$T=500$-step variance-preserving schedule as HarmoCore's diffusion
prior (Appendix~\ref{app:diff_impl}), for $200$ epochs (Adam, learning
rate $10^{-4}$, batch size $32$); inference uses DPS-style guidance
with weight $\zeta=0.3$.

\textbf{LRTFR}~\cite{luo2024lrtfr}: at test time, per-case core
coefficients are recovered by a linear least-squares fit of the
observed positions against a fixed spatial basis, then decoded to the
full grid.

\section{Additional Results and Ablations}
\label{app:results}

This appendix collects supporting quantitative and qualitative results
that complement the main paper.

\subsection{Full Quantitative Tables}
\label{app:full_tables}

Table~\ref{tab:full_10pct} extends Table~\ref{tab:main_all} to $10\%$
sensing, the ratio dropped from the main-paper table for width
(Section~\ref{sec:exp:setup}).
The 2D Synthetic benchmark has no governing PDE at any sensing ratio
(Section~\ref{sec:method:dps}), so its Phys.\ Res.\ column is fixed at
``---'' for all methods, matching Table~\ref{tab:main_all}.

\begin{table}[t]
\centering
\caption{Reconstruction Relative L2 Error and physics residual at
$10\%$ sensing (mean$\pm$std), complementing Table~\ref{tab:main_all}.}
\label{tab:full_10pct}
\begin{tabular}{lcc}
\toprule
Method & Rel.\ L2 & Phys. Res. \\
\midrule
\multicolumn{3}{l}{\textit{2D Helmholtz}} \\
Ours & \textbf{0.012$\pm$0.003} & \textbf{0.032$\pm$0.001} \\
LRTFR & 0.177$\pm$0.104 & 0.116$\pm$0.108 \\
FNO & 0.062$\pm$0.035 & 0.161$\pm$0.035 \\
F-FNO & 1.408$\pm$0.437 & 4.012$\pm$1.680 \\
VoronoiCNN & 0.116$\pm$0.064 & 0.131$\pm$0.017 \\
DiffusionPDE & 0.101$\pm$0.066 & 0.594$\pm$0.170 \\
\midrule
\multicolumn{3}{l}{\textit{2D Synthetic Wave Fields}} \\
Ours & \textbf{0.025$\pm$0.010} & --- \\
LRTFR & 0.009$\pm$0.003 & --- \\
FNO & 1.010$\pm$0.033 & --- \\
F-FNO & 1.031$\pm$0.053 & --- \\
VoronoiCNN & 0.064$\pm$0.030 & --- \\
DiffusionPDE & 0.031$\pm$0.012 & --- \\
\midrule
\multicolumn{3}{l}{\textit{3D Helmholtz}} \\
Ours & \textbf{0.131$\pm$0.105} & \textbf{5.001$\pm$2.854} \\
LRTFR & 0.499$\pm$0.473 & 372$\pm$830 \\
FNO & 1.170$\pm$0.121 & 1080$\pm$480 \\
F-FNO & 1.147$\pm$0.118 & 968$\pm$364 \\
VoronoiCNN & 0.138$\pm$0.038 & 6.496$\pm$3.642 \\
DiffusionPDE & 0.161$\pm$0.033 & 60.56$\pm$33.80 \\
\bottomrule
\end{tabular}
\end{table}

\begin{table*}[t]
\centering
\caption{Reconstruction Rel.\ L2 error (mean$\pm$std) on the 2D
Synthetic OOD test set (shifted wave-speed range, no retraining),
complementing the condensed Table~\ref{tab:ood_synth}.}
\label{tab:ood_synth_full}
\setlength{\tabcolsep}{6pt}
\begin{tabular}{lcccc}
\toprule
Method & 1\% & 2\% & 5\% & 10\% \\
\midrule
\textbf{Ours} & \textbf{0.254$\pm$0.107} & \textbf{0.097$\pm$0.052} & \textbf{0.033$\pm$0.014} & \textbf{0.025$\pm$0.011} \\
LRTFR & 0.496$\pm$0.034 & 0.206$\pm$0.024 & 0.174$\pm$0.096 & 0.112$\pm$0.037 \\
FNO & 1.119$\pm$0.040 & 1.110$\pm$0.039 & 1.083$\pm$0.037 & 1.043$\pm$0.035 \\
F-FNO & 1.087$\pm$0.062 & 1.083$\pm$0.062 & 1.069$\pm$0.064 & 1.045$\pm$0.067 \\
VoronoiCNN & 0.455$\pm$0.305 & 0.307$\pm$0.248 & 0.154$\pm$0.161 & 0.091$\pm$0.109 \\
DiffusionPDE & 0.542$\pm$0.262 & 0.326$\pm$0.290 & 0.140$\pm$0.262 & 0.093$\pm$0.215 \\
\bottomrule
\end{tabular}
\end{table*}

\subsection{Out-of-Distribution Robustness (2D Synthetic)}
\label{app:ood_synth}

To probe sensitivity to a shift in the underlying generative
distribution at test time, we build a second 2D Synthetic test set in
which the per-sample wave speed $v_s$ is redrawn from a substantially
wider range than the in-distribution setting used everywhere else in
the paper ($v_s\sim\mathcal{U}(0.8,1.2)$, empirically $v_s\in[0.84,1.19]$
across its held-out samples; Appendix~\ref{app:data}): the
out-of-distribution (OOD) test set instead draws $v_s$ from a shifted,
wider range (empirically $v_s\in[0.44,1.76]$ across its $10$ samples),
with every other generative factor---source count, source/reflector
positions and weights, the reflection coefficient $\beta$, and the
evaluation frequency grid---held fixed via the same random seed. No
model is retrained: every method uses the same checkpoint evaluated in
Table~\ref{tab:main_all}, applied unchanged to this shifted test set.
Results are reported in Table~\ref{tab:ood_synth} and discussed in
Section~\ref{sec:exp:additional}.

Table~\ref{tab:ood_synth_full} reports the same comparison in full
(mean$\pm$std, all four sensing ratios), complementing the condensed
version in Table~\ref{tab:ood_synth}, which omits standard deviations
and $10\%$ sensing for space.
Every baseline with a dedicated OOD evaluation shows increased error
under the shift, most sharply for DiffusionPDE (e.g.\ $0.038\to0.140$
at $5\%$ sensing, more than a $3\times$ increase) and, to a lesser
degree, VoronoiCNN, FNO, and F-FNO---consistent with these methods
relying on a mapping fit to the in-distribution wave-speed range.
Ours is flat to marginally lower than its in-distribution
values at every sensing ratio (e.g.\ $0.035\to0.033$ at $5\%$
sensing).

\subsection{Frequency-Conditioning Encoding for the Diffusion Prior: A Negative Result}
\label{app:freq_conditioning}

The diffusion prior conditions on frequency through the normalized
scalar $\omega_{\mathrm{norm}}\in[0,1]$ injected via FiLM at each
ResBlock (Appendix~\ref{app:diff_impl}). During development we also
tried replacing this scalar with a richer Fourier-style encoding of
$\omega_{\mathrm{norm}}$,
\begin{equation}
\begin{aligned}
\gamma(\omega_{\mathrm{norm}}) = 
&\bigl\{\sin(k\pi\,\omega_{\mathrm{norm}}),\
\cos(k\pi\,\omega_{\mathrm{norm}})\bigr\}_{k=0}^{K-1},
\end{aligned}
\label{eq:fourier_omega}
\end{equation}
with $K=8$ frequency bands, concatenated and fed through the same
FiLM conditioning path in place of the scalar default.

Table~\ref{tab:freq_conditioning} compares this Fourier encoding
against several variants---augmenting it with explicit low-order
polynomial terms, dropping the Fourier features entirely in favor of
the polynomial terms alone, further adding a linear spectral
positional term, and the scalar-only default---on 2D Helmholtz at
$10\%$ sensing. None of the richer encodings improves over the plain
scalar conditioning: the scalar-only variant attains the lowest error
(0.030), while every richer variant is worse, with no consistent
ordering among them---adding a linear spectral term on top of the
polynomial encoding gives the worst result tested (0.052), Fourier
encoding alone is only slightly better (0.049), and augmenting Fourier
features with explicit polynomial terms or using the polynomial terms
alone both land at an intermediate 0.044. We treat this as a negative
result: for this benchmark, more elaborate frequency encodings for the
diffusion prior's conditioning do not translate into improved
reconstruction accuracy, consistent with the summary in
Appendix~\ref{app:diff_impl}.

\begin{table}[t]
\centering
\caption{Reconstruction Rel.\ L2 error on 2D Helmholtz at $10\%$
sensing under different frequency-conditioning encodings for the
diffusion prior.}
\label{tab:freq_conditioning}
\begin{tabular}{lc}
\toprule
Frequency conditioning & Rel.\ L2 \\
\midrule
Fourier encoding (Eq.~\ref{eq:fourier_omega}) & 0.049 \\
Fourier + $\omega_{\mathrm{norm}}$ + $\omega_{\mathrm{norm}}^2$ & 0.044 \\
$\omega_{\mathrm{norm}}$ + $\omega_{\mathrm{norm}}^2$ (no Fourier) & 0.044 \\
$\omega_{\mathrm{norm}}$ + $\omega_{\mathrm{norm}}^2$ + linear ($k\pi\omega_{\mathrm{norm}}$) & 0.052 \\
\textbf{$\omega_{\mathrm{norm}}$ only (paper default)} & \textbf{0.030} \\
\bottomrule
\end{tabular}
\end{table}

\subsection{Additional Qualitative Comparisons}
\label{app:qual_more}

\begin{figure*}[t]
    \centering
    \includegraphics[width=0.92\textwidth]{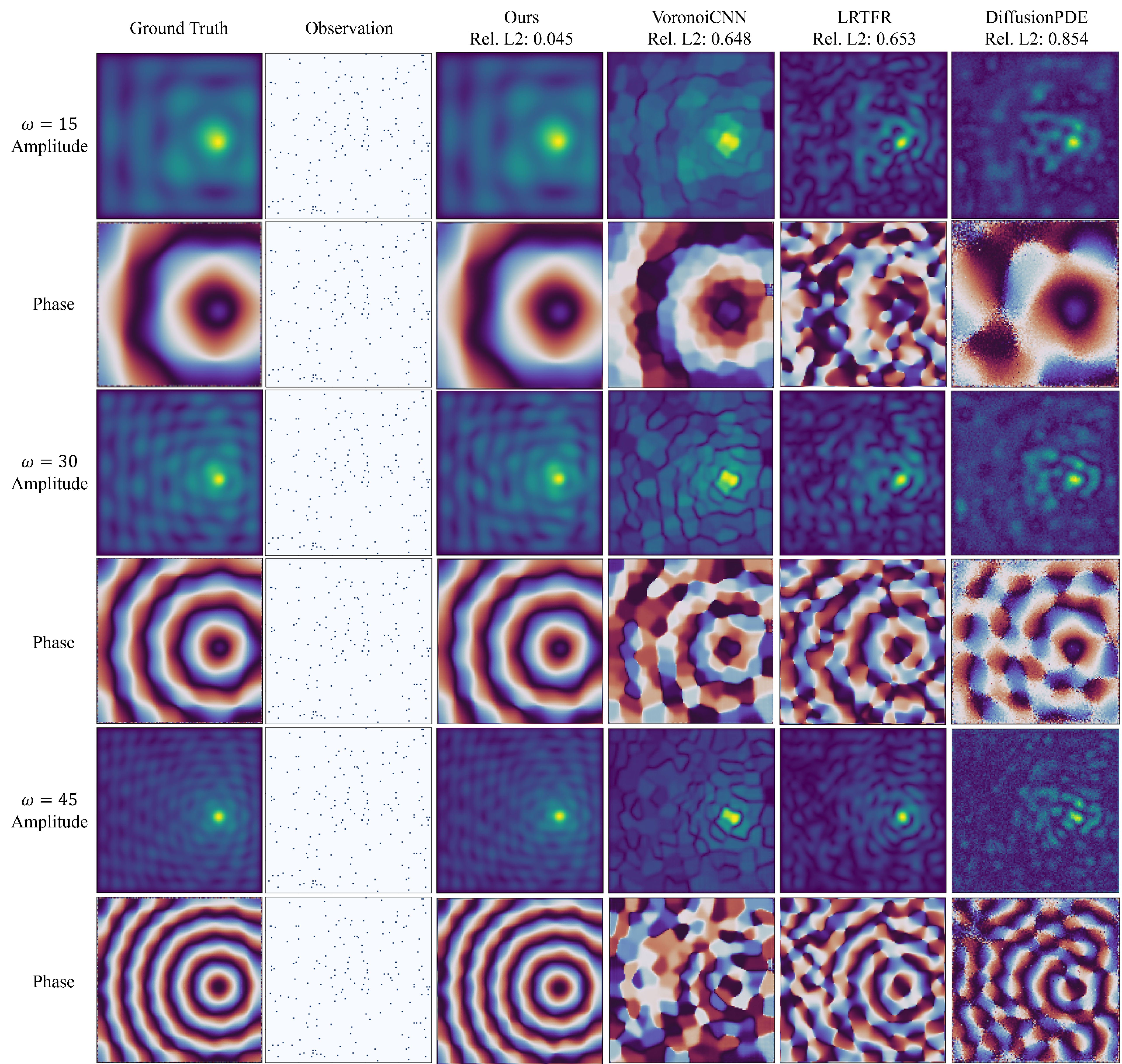}
    \caption{Additional qualitative 2D Helmholtz reconstructions across
    multiple test cases, with real and imaginary channels shown as
    separate rows per case, complementing the single-case comparison in
    Figure~\ref{fig:helmholtz2d_qual}.}
    \label{fig:helmholtz2d_qual_full}
\end{figure*}

\begin{figure*}[t]
    \centering
    \includegraphics[width=0.92\textwidth]{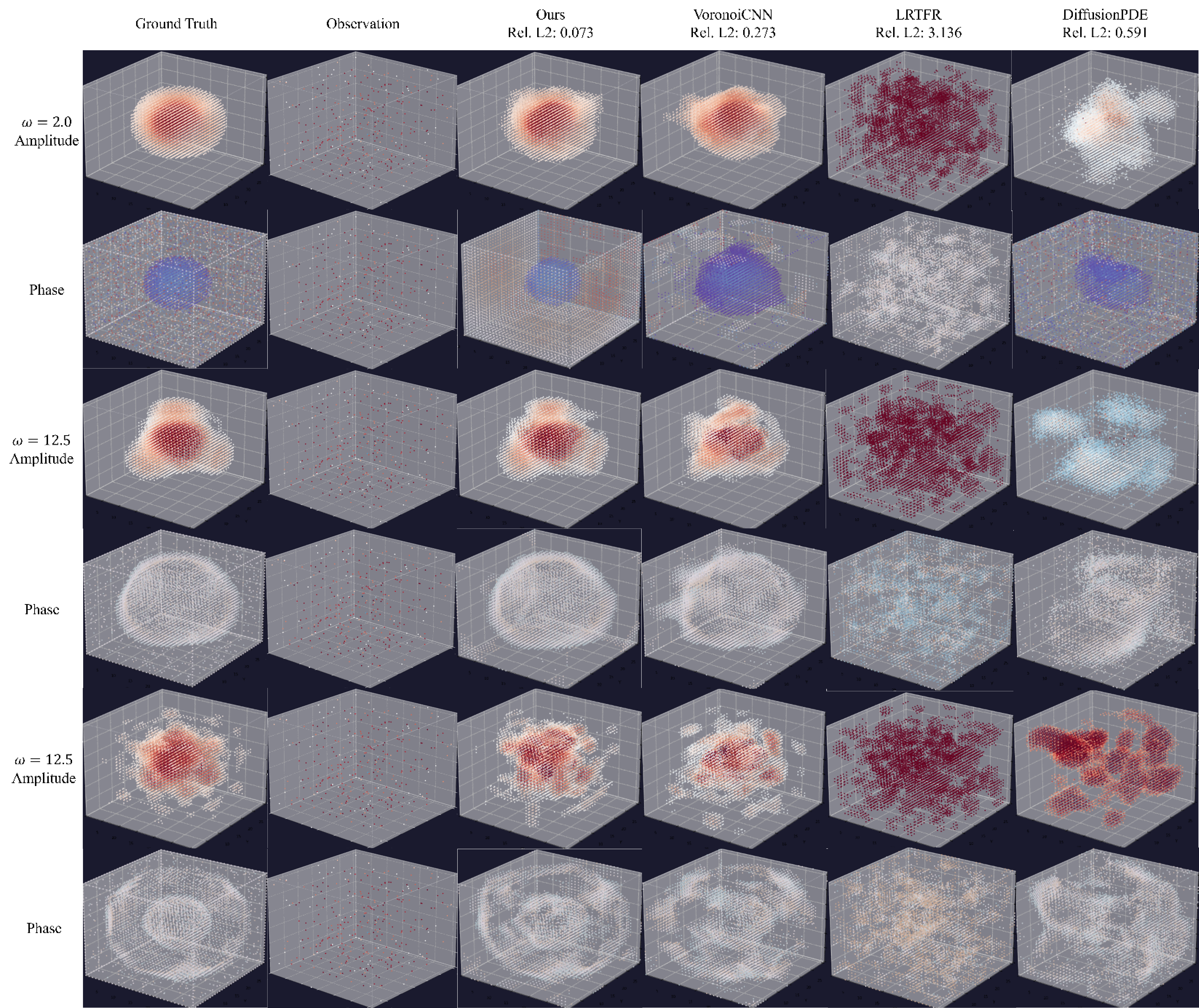}
    \caption{Additional qualitative 3D Helmholtz reconstructions across
    multiple test cases, with real and imaginary channels shown as
    separate rows per case, complementing the single-case comparison in
    Figure~\ref{fig:helmholtz3d_qual_app}.}
    \label{fig:helmholtz3d_qual_full}
\end{figure*}

\begin{figure*}[t]
    \centering
    \includegraphics[width=0.92\textwidth]{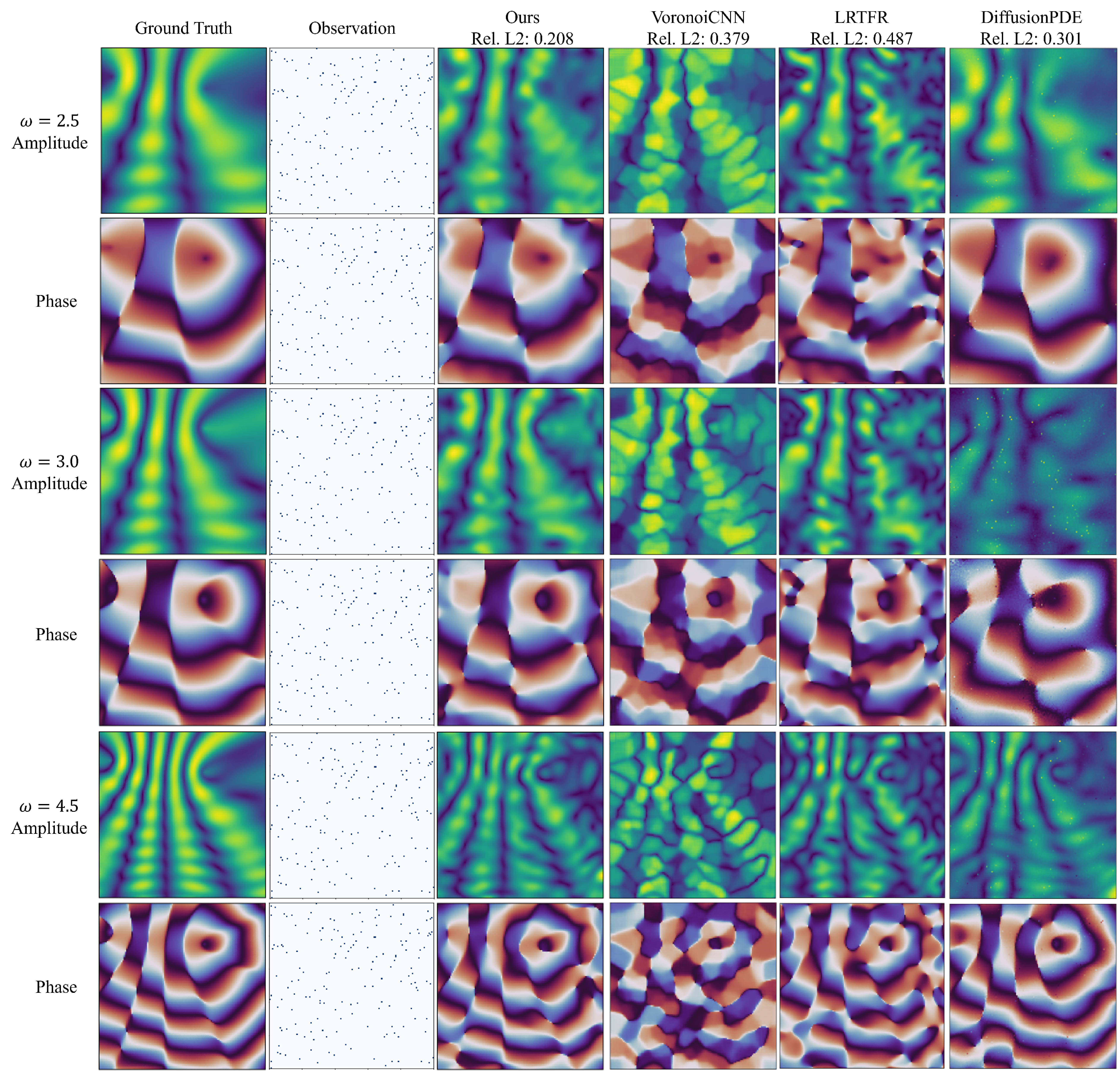}
    \caption{Qualitative 2D Synthetic wave-field reconstructions across
    three test cases, with real and imaginary channels shown as
    separate rows per case. }
    \label{fig:synthetic2d_qual_app}
\end{figure*}

Qualitative 3D Helmholtz reconstructions at $1\%$ sensing are shown in
Figure~\ref{fig:helmholtz3d_qual_app} in
Section~\ref{sec:exp:main}; Figure~\ref{fig:helmholtz2d_qual_full} and
Figure~\ref{fig:helmholtz3d_qual_full} below extend both Helmholtz
benchmarks to additional test cases.
Figure~\ref{fig:synthetic2d_qual_app} shows qualitative 2D Synthetic
wave-field reconstructions, complementing the quantitative results in
Section~\ref{sec:exp:main}; the main paper itself has no
qualitative figure for this benchmark.

\end{document}